\documentclass{article}

\PassOptionsToPackage{numbers}{natbib}
\usepackage{natbib}
\usepackage{graphicx}
\usepackage{subfigure}
\usepackage{multirow}

\usepackage[final]{neurips_2022}

\usepackage{bm}
\usepackage[utf8]{inputenc} 
\usepackage[T1]{fontenc}    
\usepackage{hyperref}       
\usepackage{url}            
\usepackage{booktabs}       
\usepackage{amsfonts}       
\usepackage{nicefrac}       
\usepackage{microtype}      
\usepackage{xcolor}         
\usepackage{wrapfig,lipsum,booktabs}

\usepackage{amsmath}
\usepackage{amssymb}
\usepackage{mathtools}
\usepackage{amsthm}
\usepackage{algorithm}
\usepackage{algorithmic}
\usepackage{longtable}
\usepackage{makecell}

\usepackage{pifont}
\newcommand{\bit}{\begin{itemize}}
\newcommand{\eit}{\end{itemize}}
\usepackage{enumitem}

\title{Region-Conditioned Orthogonal 3D U-Net \\ for Weather4Cast Competition}

\author{%
  Taehyeon Kim \quad Shinhwan Kang\quad Hyeonjeong Shin\quad Deukryeol Yoon \\
  \textbf{Seongha Eom} \quad \textbf{Kijung Shin} \quad \textbf{Se-Young Yun}\\
  KAIST AI\\
  Seoul, Korea\\
  \texttt{\{potter32, shinhwan.kang, hyeonjeong1, deukryeol.yoon\}@kaist.ac.kr} \\
  \texttt{\{doubleb, kijungs, yunseyoung\}@kaist.ac.kr} \\
}

\begin{document}

\maketitle
\begin{abstract}
The Weather4Cast competition (hosted by NeurIPS 2022) required competitors to predict super-resolution rain movies in various regions of Europe when low-resolution satellite contexts covering wider regions are given. In this paper, we show that a general baseline 3D U-Net can be significantly improved with region-conditioned layers as well as orthogonality regularizations on 1$\times$1$\times$1 convolutional layers. Additionally, we facilitate the generalization with a bag of training strategies: mixup data augmentation, self-distillation, and feature-wise linear modulation (FiLM).
Presented modifications outperform the baseline algorithms (3D U-Net) by up to $\textbf{19.54\%}$ with less than 1\% additional parameters, which won the 4th place in the core test leaderboard.
\end{abstract}

\section{Introduction}



Precipitation forecasting is one of the most arduous problem in forecasting the meteorological conditions such as air quality, solar, temperature, and wind velocity. Accurate forecasting can prevent enormous economic and social damages from a variety of applications: large-scale crop management, autonomous driving systems, and air traffic control.
While Numerical Weather Prediction (NWP) is a general method for predicting the climate changes based on the calculation of physics-based simulations, its performance for short-term rain prediction\,(i.e., less than 6 hours) is still inaccurate despite lots of computational efforts. 
Recently, deep learning techniques have attracted huge attention from the weather research community for such short-term precipitation forecasting\,\citep{ko2022effective,ko2022deep,shi2015convolutional,shi2017deep,espeholt2022deep}. Specifically, among these techniques, \citet{espeholt2022deep} develop an end-to-end deep learning method that outperforms High Resolution Rapid Refresh (HRRR)~\citep{benjamin2016north}, which is the start-of-the-art method used in United States.

Weather4Cast 2022~\citep{10.1145/3459637.3482044, 9672063} is a competition for designing the best deep learning-based precipitation forecasting model where competitors attempted to forecast super-resolution rainfall events for the next 8 hours at 15-minute intervals from low-resolution satellite radiances over various regions in Europe.
For the stage2 task in which our team participated, the desired model for competitors is to predict the 7 Europe regions across two years (2019 and 2020) when the training dataset is composed of spectral satellite imagery which covers larger areas with low-resolutions having 11 input variables for each pixel. While the satellite imagery data demands for the spatio-temporal modelling, the studies for conventional methods are still under-explored for being robust towards such \textit{spatio-temporal shifts}.






To tackle the challenge of spatio-temporal shifts, we propose a \underline{R}egion \underline{C}onditioned \underline{N}etwork (RCN) to inject the regional information into the output of 3D residual U-Net's encoder architecture, which is the variation of 3D U-Net~\citep{cciccek20163d}.
With given spectral satellite contexts of different regions, RCN can extract the region-conditioned context and such contexts linearly modulate the output of the 3D U-Net.
In addition, by penalizing the orthogonality regularization for the 1$\times$1$\times$1 convolutional layer, the network can capture more fined-grained representations for the super-resolution prediction so that it yields the better score. We also stabilize the training from the spatio-temporal shifts of the dataset.
Lastly, we adapt a bag of training strategies such as mixup~\citep{zhang2017mixup}, self-distillation, and feature-wise linear modulation (FiLM)~\citep{anonymous2023fit,perez2018film}.
More precisely, we add FiLM layers to a backbone model for fine-tuning the layers for each region of each year while freezing other backbones except the FiLM layers.
We provide more details about RCN and training strategies in Section~\ref{sec:3}.

Our contributions can be summarized as follows:

\begin{itemize}[leftmargin=9pt]
\vspace{-2mm}
\item \textbf{Effective:} We utilize two concepts: (1) region-conditioned network and (2) orthogonality regularization on 1$\times$1$\times$1 convolutional layers. With additional training strategies, our solution outperforms a baseline up to \textbf{19.54\%} with less than \textbf{1\% additional parameters}.
\item \textbf{Applicable:}  Our approaches can be adapted to any other deep neural networks.
\item \textbf{Reproducible:} We provide our source code at \cite{codeurl}.
\end{itemize}




\section{Overview of Weather4Cast Challenge and Provided Data}
\label{sec:2}
\subsection{Weather4Cast Challenge}
The main objective of Weather4Cast competition is to predict future super-resolution rainfall events (i.e., rain or no-rain) from lower-resolution satellite radiance. 
In this competition, competitors are required to provide a model predicting rainfall events until eight hours in 32 time slots for given 4 time slots of a proceeding hour. 
As the given data is composed of multiple regions in Europe across two years, a key is to learn the robust model under spatio-temporal shifts.

The challenge comprised two different tasks: (1) \textbf{stage1}: predicting 3 different regions for 1 year (2019) and (2) \textbf{stage2}: 7 different regions for 2 years (2019 and 2020). Additionally, the rain rate threshold for the latter task is 0.2 while the former one is 0.0001. The solution for the stage2 task can bring beneficial meteorological meanings while it is a harder challenge which increases the sparsity of the rain events to be predicted. In this paper, our solution targets for the stage2 task. 



\subsection{Dataset}
    The dataset is provided with satellite imagery including 11 observed physics-information, positional information, and observed rainfall amounts. The detailed explanations are as follows:




\begin{itemize}[leftmargin=9pt]
    \item \textbf{Regions:} {The dataset consists of 3 different regions for 1 year (2019) in stage1, and it is extended to 7 regions for 2 years (2019 and 2020) in stage2.}
    \item \textbf{Input variables:} Each spectral satellite imagery include 11 variables which are slightly noisy satellite radiances covering visible, water vapor, and infrared bands: IR\_016, IR\_039, IR\_087, IR\_097, IR\_108, IR\_120, IR\_134, VIS006, VIS008, WV\_062, and WV\_073. Detailed information for each context is not provided. 
    \item \textbf{Sequential information:} Each input image covers 15 minutes where each pixel corresponds to 12km $\times$ 12km area while each pixel for the output indicates 2km $\times$ 2km area.
    \item \textbf{Rainfall amount:} Pixel-wise rainfall information is provided as a float value. The precipitation ratio is in Table~\ref{tab:precipitation ratio}.
    \item \textbf{Static information:} Metadata contains the information of latitude, longitude, and height for each pixel.
\end{itemize}

\begin{wraptable}{r}{6.5cm}
\centering \footnotesize
\vspace{-40pt}
\caption{Statistics over different regions: boxi0015, boxi0034, and boxi0076.}
\begin{tabular}{@{}cccc@{}}
\toprule
Region & boxi0015    & boxi0034    & boxi0076    \\ \midrule
No-rain     & 0.810 & 0.810 & 0.892 \\
Rain     & 0.190 & 0.190 & 0.108 \\ \bottomrule
\end{tabular}
\label{tab:precipitation ratio}
\end{wraptable}

\subsection{Evaluation Metrics}
%

We evaluate the predictive performance in terms of Critical Success Index (CSI) score~\cite{schaefer1990critical}, F1-score, accuracy, and Intersection over Union (IoU). 
In particular, CSI-score is the common evaluation metric in precipitation forecasting. It is the total number of correct event forecasts divided by the sum of the total number of storm forecasts and the number of misses, i.e., 
\begin{equation}
    \textsc{CSI} = \frac{\text{TP}}{\text{TP} + \text{FN} + \text{FP}}
\end{equation}

where TP, FN, and FP are true positive, false negative, and false positive, respectively.



\begin{figure}[t]
\begin{center}
   \includegraphics[width=1.0\linewidth]{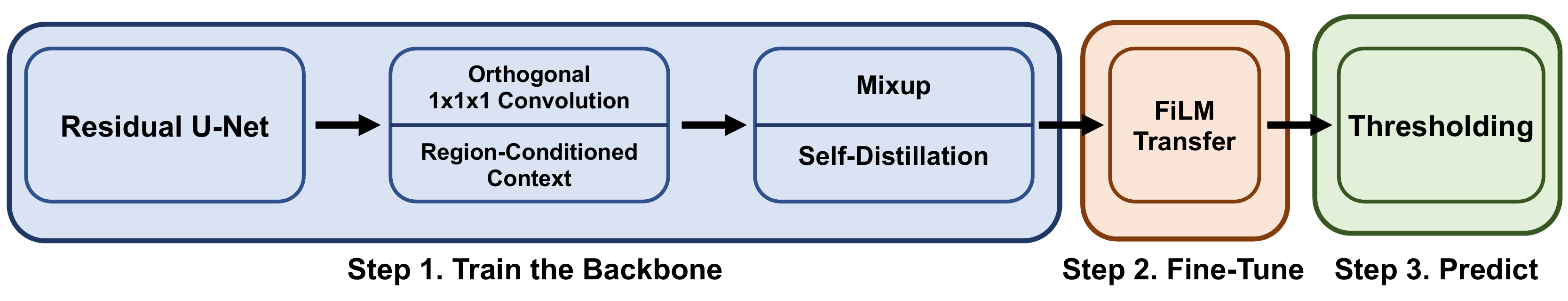}
\end{center}
\caption{An overview of our solution in the task of predicting the 7 regions at 2019 and 2020.}
\label{fig:overall_train}
\end{figure}

\section{Method}
\label{sec:3}
This section provides the solution of KAIST AI. Overall training consists of 3 steps: (1) train the backbone (Residual U-Net) with orthogonal 1$\times$1$\times$1 convolutional layers as well as region-conditioned, (2) fine-tune the backbone with FiLM Transfer approach\,\cite{anonymous2023fit} for each region of a certain year, and (3) predict the output via thresholding\,(\autoref{fig:overall_train}).

\subsection{Baseline: 3D U-Net}
We choose the baseline model as a 3D U-Net \citep{cciccek20163d}, which  utilizes the same layers with the convolutional encoder-decoder architecture for the volumetric segmentation task, on the region `boxi0015', `boxi0034', `boxi0076' in 2019 with DiceBCEloss~\cite{zhou2018unet++}. 
As \autoref{tab:baseline} shows, the baseline performance can be improved more as the batch size increases.  Interestingly, such baseline models make fairly accurate predictions for different regions even without the use of region information as an input. However, to make super-resolution predictions more accurate, conditioning for regions is needed during the propagation.

\begin{table}[h]
\centering
\caption{A preliminary survey of the 3D U-Net architecture on the validation dataset. We set the regions to `boxi0015', `boxi0034', and `boxi0076' in 2019.}
\begin{tabular}{@{}cccccccc@{}}
\toprule
Batch size & CSI-score    & F1-score     & Accuracy & IoU    & Loss   & Precision & Recall \\ \midrule
16         & 0.3130 & 0.4668 & 0.7302   & 0.3130 & 0.7670 & 0.3310    & 0.8321 \\
32         & 0.3221 & 0.4767 & 0.7499   & 0.3221 & 0.7638 & 0.3457    & 0.8043 \\
48         & 0.3303 & 0.4934 & 0.7348   & 0.3303 & 0.7629 & 0.3454    & 0.8790 \\ \bottomrule
\end{tabular}
\label{tab:baseline}
\end{table}

\begin{figure}[!t]
\begin{center}
   \includegraphics[width=1.0\linewidth]{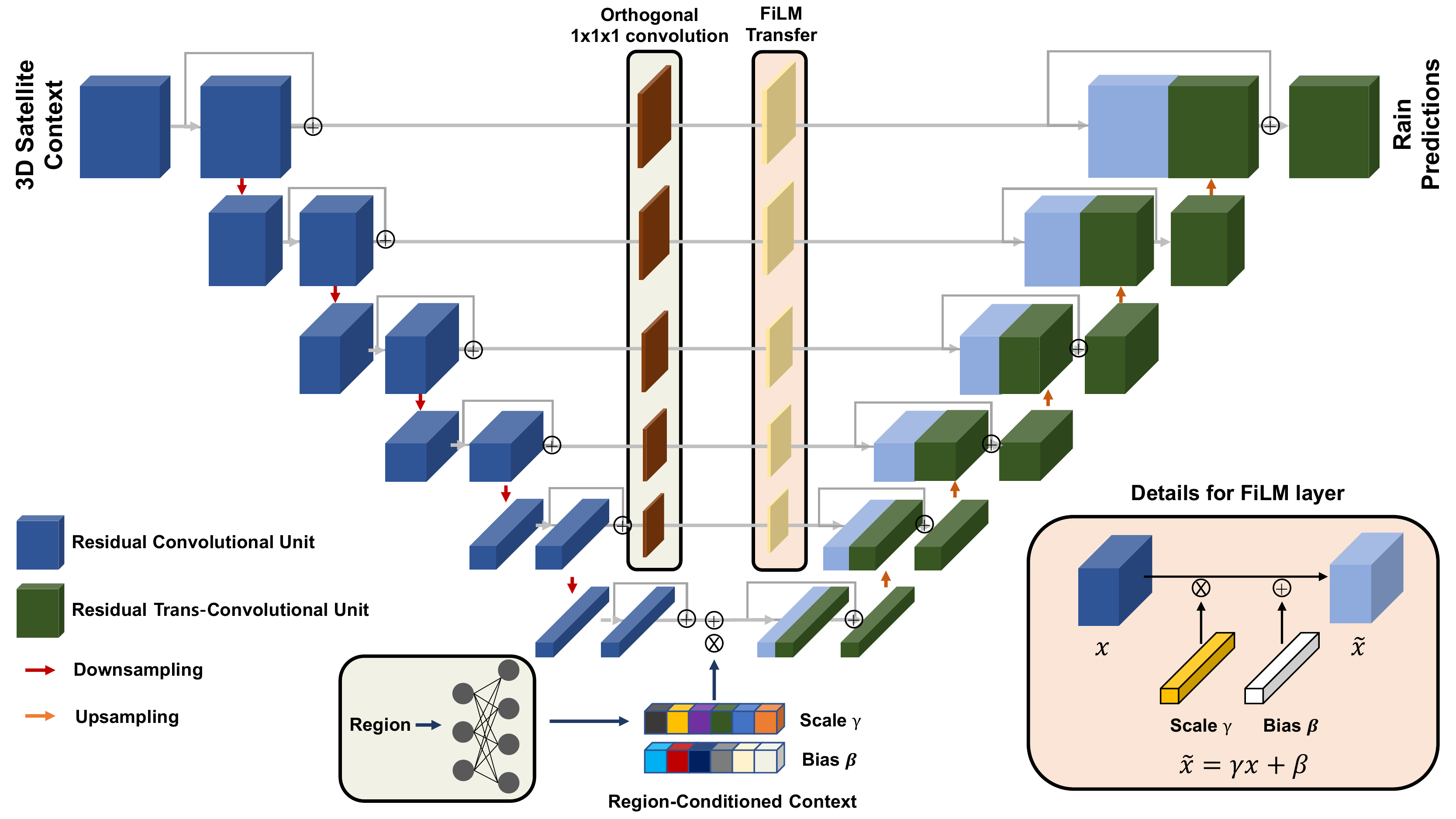}
\end{center}
\caption{An overview of our modified U-Net architecture. Each blue box corresponds to the residual convolutional unit and each green block denotes the residual transposed convolutional unit. During the propagation, the region-conditioned context is added to the last output of the encoder while the shortcut from the encoder unit to the corresponding decoder unit is transformed with orthogonal 3D 1$\times$1$\times$1 convolutional opertors as well as FiLM layer. The arrow denotes the propagation of a multi-channel feature map.}
\label{fig:overview}
\end{figure}

\subsection{Region Conditioning}
To inject region information into feature maps during the propagation, we propose a new \underline{R}egion \underline{C}onditioned \underline{N}etwork\,(RCN) to generate region-conditioned context\,(\autoref{fig:overview}). RCN is a method of adding an auxiliary region conditioner by using two layered fully-connected networks with a ReLU activation function. Here, we transform the region categorical variables into one-hot vector. Because the given dataset is comprised of satellite contexts for 7 different regions in Europe, the length of one-hot vector is 7. We extract the region-conditioned contexts including scale $\mathbf{\gamma}$ vector and bias $\mathbf{\beta}$ vector as an output of RCN with a categorical input and formulate the feature map as follows:
\begin{equation}\label{eq:region_condition}
\begin{gathered}
    \mathbf{\gamma}, \mathbf{\beta} \leftarrow \textsc{Region Conditioned Network} (\mathbf{x_r}) \\
    \mathbf{\tilde{x_r}} \leftarrow \mathbf{\gamma} \odot \mathbf{x_r} + \mathbf{\beta}
\end{gathered}
\end{equation}

\noindent where $\mathbf{x_r}$ is the last representation output of the encoder architecture. For the detailed computation of the $\mathbf{\tilde{x_r}}$, $\gamma$ is element-wisely multiplied with $\mathbf{x_r}$ in a pointwise manner\,($\odot$), and $\beta$ is added similarly.

\subsection{Orthogonal 1$\times$1$\times$1 Convolution and Residual Unit}

Orthogonal convolutional kernel is a class of the advanced normalization techniques to preserve the magnitude of the propagation signal as well as to reduce the redundant features in the filter response. Because there is the difference of resolution size between input and output, it is needed to capture more fine-grained features from the latent representations. Inspired by the orthogonal concept, we alleviate such issue by adding 1$\times$1$\times$1 convolution into the path from the encoder block to the corresponding decoder block\,(\autoref{fig:overview}) and making those 1$\times$1$\times$1 convolutions soft orthogonal with the orthogoanlity regularization, termed as Spectral Restricted Isometry Property$^+$\,(SRIP$^+$) referring to as \,\citet{9804718}, as follows:
\begin{equation}
    \frac{\lambda}{|\mathcal{W}|} \sum_{\mathbf{W} \in \mathcal{W}} \sigma(\mathbf{W^\top W} - \mathbf{I}_n)
    \label{equation5}
\end{equation}
where $\mathbf{W}$ is a weight matrix of 1$\times$1$\times$1 convolutional kernel, $\mathcal{W}$ is a set of 1$\times$1$\times$1 convolutional kernel's weight matrices, and $\sigma(\mathbf{W^\top W} - \mathbf{I}_n)$ is the output of the power method.
\begin{equation}\label{eq:app_sp}
\begin{gathered}
    u \leftarrow (\mathbf{W^\top W} -\mathbf{I}_n)v, v \leftarrow (\mathbf{W^\top W} -\mathbf{I}_n)u,\\
    \sigma(\mathbf{W^\top W} - \mathbf{I}_n) \leftarrow \frac{\|v\|}{\|u\|}.
\end{gathered}
\end{equation}

\noindent where the vector $v \in \mathbf{R}^n$ is randomly initialized with normal distribution. The key difference from \citet{9804718} is whether the dimension of the penalized weight is 5D convolution or 4D convolution. Although we can not quantitatively/qualitatively confirm the visible latent feature-map changes through orthogonality, we observe the improvement in CSI-score performance in the validation set.

\paragraph{Residual Unit.}
We design a 3D-Residual U-Net, which is a variant of the baseline 3D U-Net\,\cite{cciccek20163d}\,(\autoref{fig:overview}). The main difference between the baseline and ours is the block type. We make a shortcut for each encoder and decoder block while an 1$\times$1$\times$1 convolutional layer is added if there is the difference of number of channels between input and output.


\subsection{Data Augmentation: Mixup}
Since satellite imagery datasets rarely contain rainfall data, compared to the abundant non-rainfall data, the model is easily biased towards the majority class, i.e., non-rainfall.
To mitigate the bias on the majority class and encourage elaborated classification on the minority class, we applied Mixup~\cite{zhang2017mixup}, a popular data augmentation technique.
Mixup regularizes neural networks by utilizing the convex combination of training data without large computational overhead, and the effectiveness is proved over various image classification tasks and semantic segmentation tasks~\cite{zhang2017mixup}.
Generally, mixup is utilized on 4D datasets, i.e., ($x_i$, $y_i$) $\in \mathbb{R}^{B\times C \times H\times W}$ where $B$ is batch size, $C$ is the number of channels, $H$ is the height, and $W$ is the width, while it is under-explored on 5D dataset which time-dimension is added. We formulate the augmented training data in the same manner for the general Mixup, and the details are as follows:
\begin{equation*}
\begin{aligned}
&\tilde{x}=\lambda x_i+(1-\lambda)x_j,\\
&\tilde{y}=\lambda y_i+(1-\lambda)y_j,
\end{aligned}
\end{equation*}
where ($x_i$, $y_i$) $\in \mathbb{R}^{B\times C \times T \times H\times W}$ are training data and ground truth target, ($x_j$, $y_j$) are randomly shuffled data of ($x_i$, $y_i$), and $\lambda\sim$ Beta($\alpha$, $\alpha$).
We fixed $\alpha=1$ after exploring some values.
Interestingly, as seen in Table~\ref{tab:mixup}, the model utilizing mixup achieved the visible performance improvement in terms of F1-score, CSI-score, and IoU.

\begin{table}[h]
\centering
\caption{Comparison of baseline and model utilizing mixup on the validation dataset.}
\begin{tabular}{@{}cccccccc@{}}
\toprule
Methods & CSI-score    & F1-score     & Accuracy & IoU    & Loss   & Precision & Recall \\ \midrule
Baseline (best)         & 0.3303 & 0.4934 & 0.7348   & 0.3303 & 0.7629 & 0.3454    & 0.8790 \\
Mixup         & 0.3699 & 0.5397 & 0.7340   & 0.3699 & 0.8024 & 0.3877    & 0.8926 \\ \bottomrule
\end{tabular}
\label{tab:mixup}
\end{table}

\subsection{Self-Distillation}
After training the region-conditioned backbone, it is re-trained with self-distillation, without the supervision of ground-truth labels.
Since the ground-truth observation is sparse and noisy, its training is unstable as well as over-confident.
To release this concern, we apply the self-distillation loss which is a well-known technique for smoothing the loss landscape to lead to a flatten optima.

\subsection{Feature-wise Linear Modulation (FiLM) Layer for further Fine-Tuning}

As the last step, the self-distilled model is fine-tuned with each regional data for each year\,(i.e., we have the 7$\times$2 architectures). Because the pre-trained model can be under-performed for a specific region (or year) while it captures the general features for all regions, the pre-trained model needs further updates for personalization. For fine-tuning, we use the FiLM Transfer\,(FiT) inspired by \cite{anonymous2023fit}, which fixes the pre-trained backbone and fine-tunes only FiLM adapter layers\,(\autoref{fig:overview}). 
We initialize the new learnable parameters for linear modulation: scale $\mathbf{\gamma_f}$ and bias $\mathbf{\beta_f}$ and modify the latent representations of the shortcut path from encoder to decoder during fine-tuning:

\begin{equation}
\begin{gathered}
    \mathbf{\tilde{x_r}} \leftarrow \mathbf{\gamma_f} \odot \mathbf{x_r} + \mathbf{\beta_f}
\end{gathered}
\end{equation}

\noindent where the detailed computation of the $\mathbf{\tilde{x_r}}$ is the same with Equation~\eqref{eq:region_condition}. Here, we do not use an auxiliary network like RCN, but directly use scale and bias parameters as learnable parameters.

\subsection{Thresholding}
We generate the optimal precipitation output by controlling the threshold for each region across a year. Generally, the model decides positive rain if the corresponding probability is over 0.5. However, because all regions have different precipitation distributions, i.e., different precipitation scales, this characteristic leads to a sub-optimal due to different scales of certainty on each region, even after the FiLM Transfer. 
Thresholding approach is a ubiquitous technique to tune the prediction in a post-processing manner\,\cite{espeholt2022deep}.
More precisely, given $p \in [0,1]$, the points with probability higher than $p$ is decided as positive rain.
We explore the best threshold with fixing bin into 0.1, i.e., $p=0.1, 0.2,\cdots, 0.9$. As a result, the relaxed threshold enhances the rain generalization across several regions, and the best performance can be achieved with the combination of thresholds $p=0.1, p=0.2,$ and $p=0.4$ for different regions. Especially, `boxi0076' region, which has little precipitation observations, is significantly improved.

\subsection{Training Details and Leaderboard Results}
Table~\ref{tab:hyperparameter} includes the value of hyperparameters used in this work. Through the combination of our proposed methods, we can achieve more generalized score\,(\autoref{tab:leaderboard}).

\begin{table}[h]
\centering
\caption{Hyperparameter settings.}
\resizebox{\textwidth}{!}{
    \begin{tabular}{@{}c|cccccc@{}}
        \toprule
        Hyperparameter  & Optimizer & Learning rate & Maximum epoch & Dropout rate & Patience & Batch Size \\ \midrule
        Value &  AdamW & 1e-4 & 90 & 0.4 & 40 & 56 \\
        \bottomrule
    \end{tabular}
}
\label{tab:hyperparameter}
\end{table}


\begin{table}[h]
\centering\small
\caption{The leaderboard score (i.e., IoU) of our solution for stage2 compared to the baseline score submitted by the organizer.}
\resizebox{\textwidth}{!}{%
\begin{tabular}{@{}cccccccccc@{}}
\toprule
\multirow{2}{*}{Task} &
  Region &
  \multicolumn{2}{c}{boxi0015} &
  \multicolumn{2}{c}{boxi0034} &
  \multicolumn{2}{c}{boxi0076} &
  \multicolumn{2}{c}{roxi0004} \\ \cmidrule(lr){2-10}
                              & Year     & 2019  & 2020  & 2019  & 2020  & 2019  & 2020  & 2019  & 2020  \\ \midrule
\multirow{2}{*}{Core Test}    & Baseline & 0.223 & 0.163 & 0.149 & 0.298 & 0.268 & 0.070 & 0.204 & 0.296 \\
                              & Ours     & 0.270 & 0.237 & 0.200 & 0.362 & 0.294 & 0.104 & 0.234 & 0.321 \\ \midrule
\multirow{2}{*}{Core Heldout} & Baseline & 0.299 & 0.193 & 0.243 & 0.238 & 0.155 & 0.369 & 0.272 & 0.251 \\
                              & Ours     & 0.328 & 0.210 & 0.279 & 0.237 & 0.166 & 0.328 & 0.294 & 0.311 \\ \midrule
\multirow{2}{*}{Task} &
  Region &
  \multicolumn{2}{c}{roxi0005} &
  \multicolumn{2}{c}{roxi0006} &
  \multicolumn{2}{c}{roxi0007} &
  \multicolumn{2}{c}{\multirow{2}{*}{Overall}} \\ \cmidrule(lr){2-8}
                              & Year     & 2019  & 2020  & 2019  & 2020  & 2019  & 2020      \\ \midrule
  \multirow{2}{*}{Core Test}    & Baseline & 0.240 & 0.228 & 0.341 & 0.274 & 0.327 & 0.089 & \multicolumn{2}{c}{0.226} \\
                              & Ours     & 0.281 & 0.272 & 0.384 & 0.336 & 0.361 & 0.133 & \multicolumn{2}{c}{0.271} \\ \midrule
  \multirow{2}{*}{Core Heldout} & Baseline & 0.274 & 0.299 & 0.325 & 0.403 & 0.245 & 0.005 & \multicolumn{2}{c}{0.255} \\
                              & Ours     & 0.287 & 0.318 & 0.389 & 0.417 & 0.248 & 0.028 & \multicolumn{2}{c}{0.274} \\ \bottomrule
\end{tabular}%
}
\label{tab:leaderboard}
\end{table}

\section{Conclusion}
\label{sec:conclusion}

In this work, we propose the region-conditioned orthogonal residual U-Net model for precipitation forecasting. The performance of the proposed model outperforms the 3D U-Net model by up to $19.54\%$.
Our contributions can be folded into three perspectives.
Firstly, we renovate the original 3D U-Net with effective techniques, that are region-conditioned layers and orthogonality regularization on 1$\times$1$\times$1 convolutional layers.
Next, the generalization capability of the model can be enhanced by utilizing common training techniques such as mixup data augmentation, self-distillation, and feature-wise linear modulation. 
Lastly, our solution can be easily reproduced, and thus our repository can offer a great entry point to facilitate future developments in rainfall forecasting for data scientists.
These interesting approaches are not limited to 3D U-Net, but could be applied to other architectures designed for precipitation forecasting, such as ConvLSTM and MetNet.


\section*{Acknowledgement}
This work was supported by the Korea Meteorological Administration Research and Development Program "Development of AI techniques for Weather Forecasting" under Grant (KMA2021-00121) and Institute of Information \& communications Technology Planning \& Evaluation (IITP) grant funded by the Korea government(MSIT) (No.2019-0-00075, Artificial Intelligence Graduate School Program(KAIST)).

\bibliography{reference.bib}
\bibliographystyle{plainnat}


\end{document}